\documentclass[runningheads]{llncs}
\usepackage{graphicx}
\usepackage{amssymb}
\usepackage{amsmath}
\usepackage{hyperref}
\usepackage{booktabs}
\usepackage{cleveref}
\usepackage{caption}
\usepackage{subcaption}
\usepackage[misc]{ifsym}
\hypersetup{pdfpagemode={UseOutlines},
bookmarksopen=true,
bookmarksopenlevel=0,
hypertexnames=false,
colorlinks=true,
citecolor=magenta,
linkcolor=black,
urlcolor=black,
pdfstartview={FitV},
unicode,
breaklinks=true,
}
\begin{document}
\title{TaylorPDENet: Learning PDEs from non-grid Data}
%

\author{Paul Heinisch \and
Andrzej Dulny\orcidID{0009-0002-2990-9480}~\Letter \and
Andreas Hotho\orcidID{0000-0002-0483-5772} \and
Anna Krause\orcidID{0000-0003-1924-9183}}

\authorrunning{P. Heinisch, A. Dulny et al.}
%
\institute{University of Würzburg, Germany \\
\email{paul.heinisch@stud-mail.uni-wuerzburg.de}\\
\email{\{dulny,andreas.hotho,anna.krause\}@uni-wuerzburg.de}
}

%
%

\maketitle              
\begin{abstract}
Modeling data obtained from dynamical systems has gained attention in recent years as a challenging task for machine learning models.
Previous approaches assume the measurements to be distributed on a grid.
However, for real-world applications like weather prediction, the observations are taken from arbitrary locations within the spatial domain.
In this paper, we propose TaylorPDENet -- a novel machine learning method that is designed to overcome this challenge.
Our algorithm uses the multidimensional Taylor expansion of a dynamical system at each observation point to estimate the spatial derivatives to perform predictions.
TaylorPDENet is able to accomplish two objectives simultaneously: accurately forecast the evolution of a complex dynamical system and explicitly reconstruct the underlying differential equation describing the system.
We evaluate our model on a variety of advection-diffusion equations with different parameters and show that it performs similarly to equivalent approaches on grid-structured data while being able to process unstructured data as well.
 
\end{abstract}

\section{Introduction}
Dynamical systems like weather ~\cite{rasp2020weatherbench,bauer2015quiet}, chemical reactions~\cite{zhang2020data} or wave propagation~\cite{karlbauer2019distributed} are an essential part of our environment.
Analyzing these systems may lead to knowledge of how they evolve and a better understanding of the system itself \cite{raol2004modelling,close2001modeling}.
Dynamic systems are described by ODEs (ordinary differential equations) with only time derivatives or PDEs (partial differential equations) containing time and spatial derivatives \cite{kuehn2019pde}.
These equations play a vital role in many disciplines and describe the physical laws governing the system.

Modeling dynamic systems has gained attention in recent contributions as an interesting and challenging topic \cite{karlbauer2019distributed,praditia2021finite,li2020robust,so2021differential,long2019pde}.
Dynamic systems are based on the task of learning the dynamics of an underlying complex system from sequential data, to finally be able to predict future states.

\begin{figure}[ht]
\centering
\includegraphics[width=1.0\textwidth]{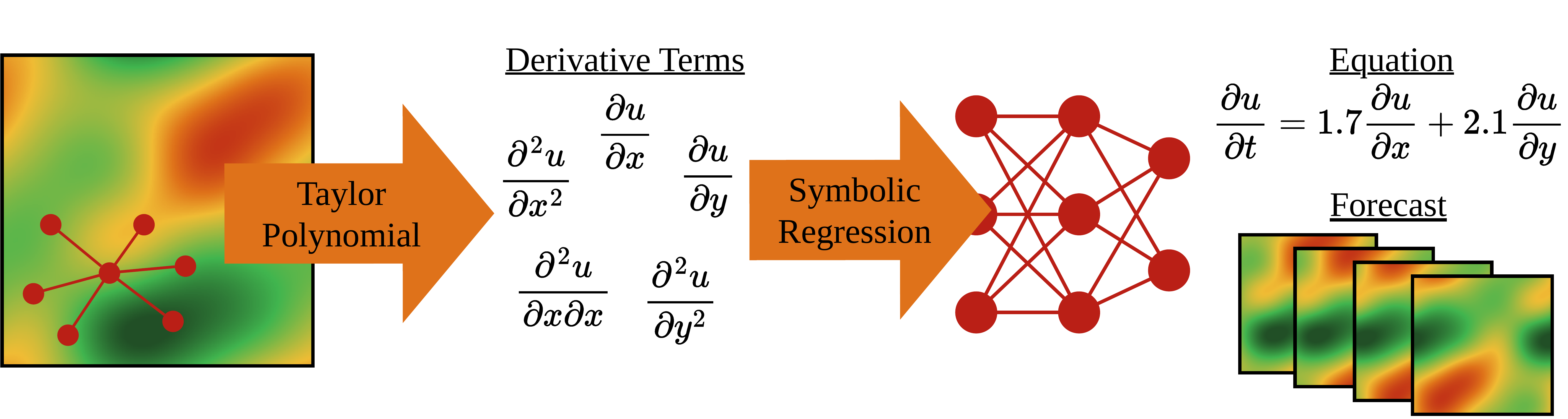}
\caption{In TaylorPDENet measurements of a system at arbitrary points in space are used with a Taylor polynomial to estimate spatial derivatives for each point. These are used with a symbolic regression network to estimate the parameters of a partial differential equation and perform predictions of future states.} \label{fig:overview}
\end{figure}

Available machine learning models require the observations to be structured on a grid.
In this case, convolutional neural networks, dominant for this type of data~\cite{long2019pde,Goodfellow-et-al-2016}, can be used to model the state of the system.
However, in real-world applications, the state of a dynamical system at a specific time is measured at arbitrary points in space.
Data, like weather dynamics \cite{rasp2020weatherbench,bauer2015quiet}, chemical reactions \cite{zhang2020data}, and wave propagation \cite{karlbauer2019distributed}, can rarely be measured on a grid.

Our objective is to develop a machine learning method able to learn from non-grid data as we often measure it in reality.
To this end, we propose the TaylorPDENet, which is based on the approximation of spatial derivatives with the Taylor polynomial \cite{gulsu2006taylor}.
TaylorPDENet does not require gridded data and instead computes the spatial derivatives $\frac{\partial^{(q+r)}u}{\partial x^q \partial y^r}$ using arbitrary neighboring points.
Our approach, as summarized in~\Cref{fig:overview}, consists of two steps:
Firstly, the spatial derivatives are calculated using the Taylor polynomial evaluated at several neighboring points.
Secondly, the differential equation is constructed using a linear symbolic regressor which assigns coefficients to different terms of the equation.
By learning the coefficients of each spatial derivative in the equation, our model is able to both accurately forecast future states of the system\cite{dulny2022neuralpde,cai2021physics} and reconstruct the underlying differential equation.

We evaluate our model on several 2D advection-diffusion equations \cite{dehghan2004numerical}:

\begin{equation}\label{eq:form}
    \frac{\partial u}{\partial t} = \alpha_{10}\frac{\partial u}{\partial x}+ \alpha_{01}\frac{\partial u}{\partial y}+ \alpha_{20}\frac{\partial^2 u}{\partial x^2}+ \alpha_{02}\frac{\partial^2 u}{\partial y^2}
\end{equation}
with different coefficients $\alpha_{ij}\in\mathbb{R}$.

Our contributions can be summarized as follows:
\begin{enumerate}
    \item we propose a novel approach to learning dynamical systems from data based on Taylor polynomials
    \item we show that our model TaylorPDENet is able to perform well on both grid and non-grid data alike
    \item we demonstrate that TaylorPDENet is able to both predict future states of the dynamical system and extract its governing equation
\end{enumerate}
We make the code containing our model and all experiments publicly available.~\footnote{The code is available at \url{https://anonymous.4open.science/r/TaylorPDENet-BAC6}}
\section{Related Work}
Previous work on learning dynamical systems from observations has mainly focused on grid-structured data.
Dulny et al.~\cite{dulny2022neuralpde} propose the NeuralPDE model which combines CNNs with a differentiable method of lines solver to parametrize the underlying PDEs.
Similarly, Ayed et al.~\cite{Ayed2019} propose a hidden-state neural-based model to forecast dynamical systems using a ResNet to parametrize the equation.
The Finite Volume Neural Network (FINN)~\cite{praditia2021finite} is based on the Finite Volume Method and predicts the evolution of diffusion-type systems by explicitly modeling the flow between grid points.
Li et al.~\cite{Li2021} propose the Fourier Neural Operator which learns the simulations of physical processing using convolutions in Fourier space.

Another line of research focuses on extracting the differential equation which describes the evolution of the dynamical system.
Long et al.~\cite{long2018pde} propose the PDE-Net model which uses learnable convolutional filters to estimate single derivative terms together with a linear regression layer to reconstruct the coefficients of the equation.
An extension of this approach is the PDE-Net 2.0~\cite{long2019pde} which additionally features a symbolic regression network capable of including non-linear terms in the equations.
Raissi et al.~\cite{Raissi2017} propose a physics-informed deep learning approach that is able to fit the coefficients of a known equation type using automatic differentiation.

The task of predicting the evolution of a physical system from non-grid structured data has only recently started gaining attention.
Recently Dulny et al.~\cite{dulny2023dynabench} proposed a benchmark for learning dynamical systems from low-resolution non-grid observations on which several graph neural network and point cloud~\cite{zhao2021point} based models were evaluated.
Iakovlev et al.~\cite{iakovlev2020learning} use a graph message passing approach to learn predictions for an advection-diffusion problem, as well as the heat equation and Burger's equation from data.
The multipole graph neural operator proposed by Li et al.~\cite{Li2020} can also be used to learn dynamical systems from unstructured data, however, the authors only evaluate it on grid observations.

To the best of our knowledge, our proposed approach TaylorPDENet is the first model capable of simultaneously forecasting the evolution of a dynamical system while also extracting the differential equations describing its behavior from non-grid observations.

\section{TaylorPDENet}
Given measurements of an evolving physical system $u\colon \Omega\times T\rightarrow \mathbb{R}^d$ at specific locations $\mathbf{p_1}, \ldots \mathbf{p_n}$ within the spatial domain $\Omega \subset \mathbb{R}^2$.
The measurements are available at time points $T = \{t_1, t_2, t_3,\dots, t_{\tau}\}$ for $T\subset \mathbb{R}$.
In principle, our approach works with arbitrary spaced time steps, but for simplicity, we assume $t_{i+1}-t_i=\text{const}$.
We denote the known state of the system at a given time $t$ as $u(t) := [u(\mathbf{p_1}, t), \ldots, u(\mathbf{p_n}, t)]\in \mathbb{R}^{n\times d}$.
The system is assumed to evolve according to an underlying PDE of the form in equation~\ref{eq:form}.
We consider the task of predicting the evolution of the system (\textit{forecasting}) as well as reconstructing the underlying differential equation (\textit{reconstruction}).

Our proposed model, TaylorPDENet, is based on the Taylor polynomial \cite{sezer2005taylor}, which offers an approximation for functions at a single point in space. 
We calculate the polynomial for the function $u(\cdot, t_i)$ at each point $\mathbf{p_i}$ and its neighboring points $\mathcal{N}(\mathbf{p_i})$ and solve a set of linear equations with the derivatives of the underlying PDE model as unknowns \cite{gulsu2006taylor}.
We combine this knowledge with learnable parameters to conduct an Euler step \cite{biswas2013discussion} and compute the future state of the model.

\subsection{Taylor Polynomial Approximation} \label{taylor_approx}
For a given point $\mathbf{p_0}\in\mathbb{R}^2$ with coordinates $x_0$ and $y_0$ in our measurements $u(t_i)$ we compute the derivatives at $\mathbf{p_0}$ using the Taylor approximation.
\begin{theorem}[Taylor Approximation~\cite{duistermaat2004multidimensional}]
\label{th:taylor_approx}
A function of two variables $f$ whose partial derivatives all exist up to the $Q^{th}$ order within a neighborhood $U$ of the point $(x_0,y_0)$ can be approximated by the \textbf{$\mathbf{Q^{th}}$-degree Taylor polynomial} of $f$ at the point $(x,y)$ in the neighbourhood of $(x_0,y_0)$ as follows:
\begin{equation}
    f(x,y) = f(x_0, y_0) + P_Q(x,y) + o(h^{Q+1})
\end{equation}
where 
\begin{equation}
    P_Q(x,y) = \sum_{q=0}^Q \sum_{r=0}^{Q-q} \frac{(x-x_0)^q(y-y_0)^r }{q!r!} \cdot \frac{\partial^{(q+r)}f}{\partial x^q \partial y^r}(x_0,y_0)
\end{equation}
and $h = \sqrt{(x-x_0)^2+(y-y_0)^2}$.
\end{theorem}
We can leverage~\Cref{th:taylor_approx} to calculate the partial derivatives of a function $f$ at a given point $\mathbf{p_0}$ using available data at $K$ neighboring points $\mathbf{p_1}, \ldots \mathbf{p_K},\mathbf{p_i}=(x_i,y_i)$.
By expanding the function at the neighboring points $\mathbf{p_i}$ using the Taylor approximation and treating the partial derivatives $u_{x^qy^r}=\frac{\partial^{(q+r)}f}{\partial x^q \partial y^r}$ as unknowns we arrive at $K$ different linear equations of the form:
\begin{equation}
    f(x_i, y_i)-f(x_0, y_0) = \sum_{q+r\leq Q; q,r\geq 0}\alpha^{(i)}(q, r)u_{x^qy^r}
\end{equation}
with $\alpha^{(i)}(q, r) = \frac{(x_i-x_0)^q(y_i-y_0)^r }{q!r!}$.

Note that the left-hand side of the equation and the coefficients $\alpha(q, r)$ can be computed with the available data.
For the the physical system $u(t)$ as the function $f$ at a given point in time $t$ and several neighboring points $\mathbf{p_i}$, we construct and solve the following linear equation system:
\begin{equation}
\label{eq:derivative_linear_system}
    \Delta \mathbf{u}=\mathbf{A}\mathbf{D}
\end{equation}
where $\mathbf{A} \in \mathbb{R}^{K\times K}$ contains all the computed factors $\alpha^{(i)}(q,r)$, $\mathbf{D} \in \mathbb{R}^K$ containing the derivatives $u_{x^qy^r}$  we want to solve for, and $\Delta \mathbf{u} = [u(t)(x_1,y_1) - u(t)(x_0,y_0),\ldots,u(t)(x_K,y_K) - u(t)(x_0,y_0)]\in \mathbb{R}^K$. 
Note that the method of computation for the derivatives $\mathbf{D}$ does not require the data points to be structured grid-like.

To calculate all partial derivatives $u_{x^qy^r}$ up to order $Q$, the linear system~\Cref{eq:derivative_linear_system} needs to contain as many equations as there are partial derivatives ($\frac{(Q+1)(Q+2)}{2}$). 
However, in practice, we find that over-specifying the equation system leads to numerically more stable results (cf.~\Cref{sec:ablation_study}).
For this reason, we use an ordinary least squares solver to estimate the derivatives $\widehat{\mathbf{D}}$ for a given point $\mathbf{p_0}$:

\begin{equation}
    \widehat{\mathbf{D}} = \underset{\mathbf{D} \in \mathbb{R}^K}{\mathrm{min}} || \mathbf{AD} -\Delta \mathbf{u} ||_2
\end{equation}
where $||\cdot||_2$ is the $l_2$ norm.

\subsection{Architecture TaylorPDENet}
We predict the state of the system one step $\Delta t$ into the future by performing one forward Euler step:

Let $u(t)$ be the state of the system at time $t$. 
Then we predict the state at time $t+ \Delta t$ as:
\begin{equation}\label{eq:prediction-step}
    \hat{u}(t+\Delta t) = w_{0, \:0}\cdot u(t) + \Delta t \cdot \sum_{q=0}^{Q} \sum_{r=0}^{Q-q} w_{q,\: r}\cdot \frac{\partial^{(q+r)}}{\partial x^q \partial y^r}u(t) \quad (j+k \neq 0)
\end{equation}
where the derivatives are estimated with the Taylor approximation at each point.
The learnable weights are denoted as $w_{q,\: r}$ with $q$ associated with the order of the derivative in $x$, and $r$ the derivative in $y$.
The weights thus represent the parameters of the underlying PDE model and can be easily extracted.

To enable the algorithm to make long-term predictions we perform several prediction steps sequentially using the output of the previous step as input for the current.
For all prediction steps the weights are shared thus keeping the coefficients constant during prediction.
Given the input data $u(t)$ we update the parameters $w_{q,\: r}$ by minimizing the error of the $L$-prediction steps  $\frac{1}{L} \sum_{i=1}^L \left(u(t+i\cdot\Delta t) - \hat{u}(t+i\cdot\Delta t) \right)^2$.
%

%
\begin{figure}[!t]
    \centering
    \includegraphics[width=\textwidth]{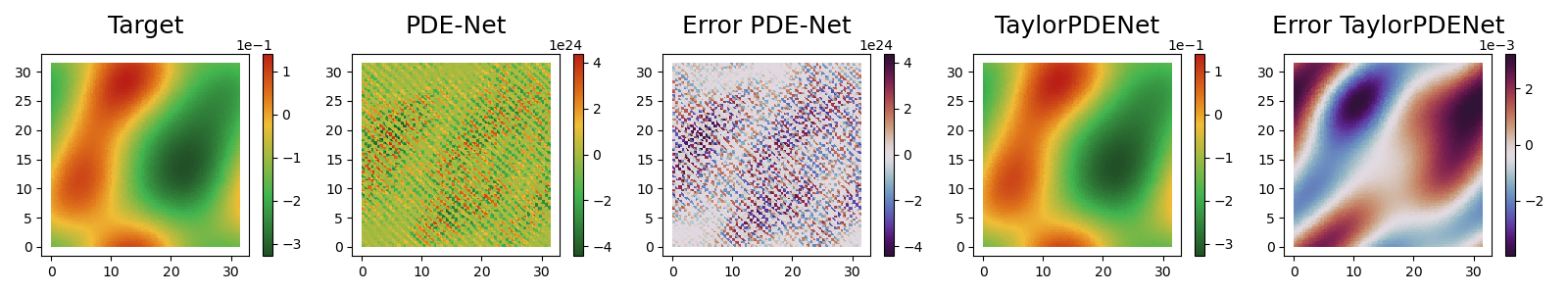}
    \caption{Comparison of the forecast done with the PDE-Net and TaylorPDENet after $L=150$ prediction steps. The forecast of the PDE-Net contains artifacts and the MSE (Mean Squared Error) is several orders of magnitude higher.}
    \label{fig:prediction_comparison}
\end{figure}
\section{Experiments}
In this section, we test the TaylorPDENets ability on grid and non-grid data, and study the influence of the number of neighbors used in the computation of the derivatives.
For grid data, we further compare the TaylorPDENet to the existing PDE-Net 2.0~\cite{long2019pde,long2018pde} as the only state-of-the-art model capable of both reconstructing the equation and forecasting the system.
For non-grid data, we only report the results of our model calculated on the same equations and number of observation points, as there are no suitable baselines for this task.
\begin{table}[!t]
\centering
\caption{Prediction MSE after $150$ prediction steps compared with the PDE-Net 2.0 evaluated on equation (3) (\Cref{tab:param_g}) with a previously unseen initial condition.
The PDE-Net diverges during forecasting beyond the number of prediction steps used during training.}
\label{tab:comparison}
\begin{tabular}{lc@{\hskip 0.15in}c@{\hskip 0.15in}c}
\toprule
Model                           & Steps & Reconstruction        & Forecast\\
\midrule
PDE-Net 2.0                     & 20    & $0.39$    & $inf$\\
TaylorPDENet (ours)             & 20    & $2.3\cdot 10^{-3}$    & $1.7\cdot10^{-6}$\\
TaylorPDENet (ours, non-grid)   & 20    & $1.8\cdot 10^{-3}$    & $4.8\cdot10^{-6}$\\
\bottomrule
\end{tabular}
\end{table}
\begin{table}[!t]
\caption{Learned PDE parameters for equation containing only first-, second-, or mixed order derivatives. The data is structured on a grid $64\times 64$ to make it comparable with the PDE-Net 2.0}\label{tab:param_g}
\centering
\begin{tabular}{l@{\hskip 0.1in}lrrrrr}
\toprule
Equation (1) &  $u_{t} =$ & $+1.500 u_{x}$ & $+1.500 u_{y}$ & & & \\
PDENet 2.0   &  $u_{t} =$ & $+0.275 u_{x}$ & $+0.288 u_{y}$ & $+0.375 u_{xx}$ & $+0.379 u_{yy}$ & $+0.378 u_{xy}$\\
TaylorPDENet &  $u_{t} =$ & $+1.518 u_{x}$ & $+1.527 u_{y}$ & $+0.121 u_{xx}$ & $+0.114 u_{yy}$ & $+0.250 u_{xy}$\\
\midrule
Equation (2) &  $u_{t} =$ & & & $+0.900 u_{xx}$ & $+0.330 u_{yy}$ & $+0.770 u_{xy}$\\
PDENet 2.0   &  $u_{t} =$ & $-0.155 u_{x}$ & $+0.132 u_{y}$ & $+0.269 u_{xx}$ & $+0.097 u_{yy}$ & $+0.267 u_{xy}$\\
TaylorPDENet &  $u_{t} =$ & $+0.000 u_{x}$ & $+0.000 u_{y}$ & $+0.903 u_{xx}$ & $+0.332 u_{yy}$ & $+0.776 u_{xy}$\\
\midrule
Equation (3) &  $u_{t} =$ & $+1.000 u_{x}$ & $+1.000 u_{y}$ & $+1.000 u_{xx}$ & $+1.000 u_{yy}$ & $+1.000 u_{xy}$\\
PDENet 2.0   &  $u_{t} =$ & $+0.143 u_{x}$ & $+0.164 u_{y}$ & $+0.231 u_{xx}$ & $+0.227 u_{yy}$ & $+0.226 u_{xy}$\\
TaylorPDENet &  $u_{t} =$ & $+1.003 u_{x}$ & $+1.005 u_{y}$ & $+1.054 u_{xx}$ & $+1.057 u_{yy}$ & $+1.108 u_{xy}$\\
\bottomrule
\end{tabular}
\end{table}
\subsection{Data}
\begin{figure}[!t]
    \centering
    \begin{subfigure}{0.22\textwidth}
        \centering
        \includegraphics[width=\textwidth]{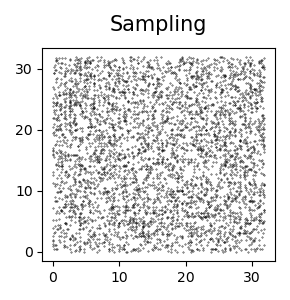}
    \end{subfigure}
    \hfill
    \begin{subfigure}{0.77\textwidth}
        \centering
        \includegraphics[width=\textwidth]{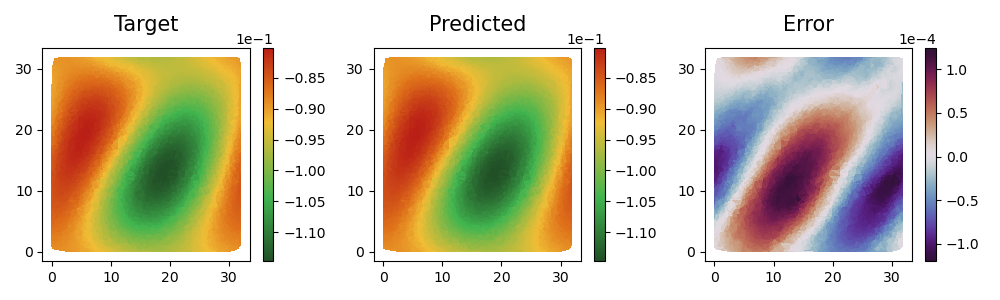}
    \end{subfigure}
    \caption{Forecast after $L=150$ prediction steps for a PDE learned from irregularly spaced data points.}
\label{fig:forecast_nongrid}
\end{figure}

For our experiments, we generate data by numerically solving a PDE for a given initial condition with the numerical simulations framework dedalus \cite{burns2020dedalus}.
The data is simulated on a grid with periodic boundary conditions of the grid size $64 \times 64$.
For our experiments on non-grid data, we use $256 \times 256$ points, which we later downsample to $64^2$ points (see \Cref{fig:forecast_nongrid}).
The solution is saved with a time discretization of $\Delta t=0.1$.
For our purposes, we generate three types of equations (see \Cref{tab:param_g,tab:param_ng}) with only first- (advection), only second- (diffusion), and all first- and second-order derivatives (advection-diffusion).
\begin{table}[!t]
\caption{Learned PDE parameter for irregularly spaced data points.}\label{tab:param_ng}
\centering
\begin{tabular}{l@{\hskip 0.1in}lrrrrr}
\toprule
Equation (1) & $u_{t} =$ & $+1.500 u_{x}$ & $+1.500 u_{y}$ &  &  & \\
TaylorPDENet & $u_{t} =$ & $+1.520 u_{x}$ & $+1.522 u_{y}$ & $+0.126 u_{xx}$ & $+0.113 u_{yy}$ & $+0.244 u_{xy}$\\
\midrule
Equation (2) & $u_{t} =$ &  &  & $+0.900 u_{xx}$ & $+0.330 u_{yy}$ & $+0.770 u_{xy}$\\
TaylorPDENet & $u_{t} =$ & $+0.000 u_{x}$ & $+0.000 u_{y}$ & $+0.904 u_{xx}$ & $+0.333 u_{yy}$ & $+0.771 u_{xy}$\\
\midrule
Equation (3) & $u_{t} =$ & $+1.000 u_{x}$ & $+1.000 u_{y}$ & $+ 1.000 u_{xx}$ & $+1.000 u_{yy}$ & $+1.000 u_{xy}$\\
TaylorPDENet & $u_{t} =$ & $+1.002 u_{x}$ & $+1.001 u_{y}$ & $+ 1.054 u_{xx}$ & $+1.053 u_{yy}$ & $+1.092 u_{xy}$\\
\bottomrule
\end{tabular}
\end{table}
\subsection{Training and Testing}\label{sec:train-test}
We use optuna \cite{akiba2019optuna} to optimize hyperparameters for the TaylorPDENet as well as the PDE-Net 2.0 \cite{long2019pde}.
During training, we use $L=20$ prediction steps and set the maximum order of the derivatives to two. 
While we use a kernel size of $5\times 5$ for the PDE-Net 2.0, we use $24$ neighbors for the TaylorPDENet. 
This way both PDE-Net and TaylorPDENet get access to the same number of neighboring points in the grid.
We evaluate the models on their performance on the forecasting and reconstruction task by computing the MSE (Mean Squared Error) between the computed and the correct values.

%
\subsection{Ablation study}
\label{sec:ablation_study}
We further investigate the influence of the number of neighbors ($K \in \{ 5, 50\}$). As was described in \Cref{taylor_approx}.
this overspecifies the equation system and increases the available data to compute the derivatives.
We use equation (3) from \Cref{tab:param_ng} with $4096$ points and $L=20$ prediction steps during training and evaluate the same way as in \Cref{sec:train-test}.
\begin{figure}
    \centering
    \includegraphics[width=0.8\textwidth]{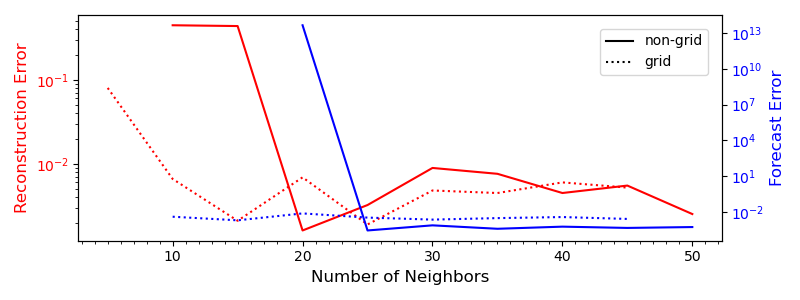}
    \caption{Reconstruction MSE (red) and forecast MSE (blue) in relation to the number of neighbors ($K$) on grid and non-grid data.}
    \label{fig:ablation}
\end{figure}
\subsection{Results}
We have shown the effectiveness of the TaylorPDENet especially on non-grid data as can be seen in \Cref{fig:forecast_nongrid}.
Moreover, the TaylorPDENet works on grid data as well and is able to reconstruct the equation and accurately predict future states more than $7\times$ further into the future than during testing as can be seen in \Cref{fig:prediction_comparison}.
Although we were not able to reproduce any results achieved with the PDE-Net 2.0 in \cite{long2019pde} on our dataset, we compare the TaylorPDENet with the PDE-Net 2.0 on grid data in \Cref{tab:comparison}.
The latter diverges during testing (\Cref{fig:prediction_comparison}) and is not able to reconstruct the equation.
The TaylorPDENet on the other hand reconstructs the equation and performs forecasting for $L=150$ prediction steps with an MSE of $4.8\cdot 10^{-6}$ on non-grid data and with $1.7\cdot 10^{-6}$ a slightly better performance on grid data.
We hypothesize the evenly distributed data points make it easier to estimate the next step.
The error for the reconstruction task is for both types of data in the same order of magnitude.
Results for the reconstruction can be seen in \Cref{tab:param_g}.
The TaylorPDENet extracts coefficients that resemble the true coefficients but we also note deviations, especially for the coefficients for the second-order derivatives that are zero in the original equation (see Equation (1) in \Cref{tab:param_g}).
We also observe a similar tendency of the TaylorPDENet on non-grid data (see \Cref{tab:param_ng}).

In the ablation study (\Cref{fig:ablation}) we found that once enough $K$-neighbors were considered, the forecast MSE does not improve significantly.
It can be seen that in general fewer neighbors are needed for grid data.
We hypothesize this is due to the evenly distributed points on grid data.
Moreover, the exactly specified system with $K=5$ does not work as expected because of linear dependencies in the system of equations.
%
\section{Conclusion and Future Work}
In this paper, we have proposed a novel architecture for learning dynamical systems from non-grid data capable of both forecasting the evolution of the system as well as reconstructing the governing equation.
We have shown in several experiments on linear advection-diffusion equations, that TaylorPDENet is capable of accurately solving both tasks on both grid and non-grid observations.
For gridded data, our model outperforms the other models capable of forecasting and equation reconstruction.
In future work, we intend to extend the approach to other types of equations.
Furthermore, we want to study the influence of noise and increase the stability and accuracy of the TaylorPDENet.
%
%
%
\clearpage
\bibliographystyle{splncs04}
\bibliography{bibliography}
\end{document}


\title{Supplementary material for TaylorPDENet: Learning PDEs from non-grid Data}
%
%


\author{Anonymous Author}
\authorrunning{Anonymous}
%
\institute{***}

\maketitle              
\appendix
\section{Hyperparameters}
The hyperparameter study was conducted with Optuna \cite{akiba2019optuna} and led to the following results. During the study, the models were trained with $L=20$ prediction steps, on a grid of $64 \times 64$ or on non-grid data $64 ^2$ downsampled points, and $10$ epochs. For the TaylorPDENet $16$ neighbors were used. The study was conducted with the TPESampler and 100 trials.
\begin{center}
\begin{tblr}{m{55pt}@{\hskip 0.5in}l}
    \textbf{ \\ TaylorPDENet \\ (non-grid)} & 
        \begin{tabular}{l@{\hskip 0.4in}l@{\hskip 0.4in}l}
        \textbf{Parameter} & \textbf{Result} & \textbf{Search space} \\
        \\
        Weight Decay & $1e^{-7}$ & log$[1e^{-3}, 1e^{-8}]$\\
        Learning Rate & $1e^{-3}$ & log$[1e^{-2}, 1e^{-4}]$\\
        Batch Size & $9$ & $[1, 16]$\\
        Neighbors & $20$ & $[10, 30]$\\
        \end{tabular} \\
        \hline
    \textbf{TaylorPDENet \\ (grid)} & 
        \begin{tabular}{l@{\hskip 0.4in}l@{\hskip 0.57in}l}
        Weight Decay & $6e^{-5}$ & log$[1e^{-3}, 1e^{-8}]$\\
        Learning Rate & $8e^{-3}$ & log$[1e^{-2}, 1e^{-4}]$\\
        Batch Size & $9$ & $[1, 16]$\\
        Neighbors & $28$ & $[10, 30]$\\
        \end{tabular}\\
        \hline
    \textbf{PDE-Net} & 
        \begin{tabular}{l@{\hskip 0.4in}l@{\hskip 0.35in}l}
        Weight Decay & $1e^{-6}$ & log$[1e^{-3}, 1e^{-8}]$\\
        Learning Rate & $1e^{-3}$ & log$[1e^{-2}, 1e^{-4}]$\\
        Batch Size & $2$ & $[1, 16]$\\
        Constraint & $moment$ & $[moment, frozen]$\\
        Kernel Size & $5$ & $[3, 5, 7]$\\
        \end{tabular}
\end{tblr}\\
\end{center}
In the table above the hyperparameters can be found as well with their search spaces. The $log$ in front of the search space indicates a logarithmic search space.
These hyperparameters were used for the most part during the experiments, except when stated otherwise.

\section{Additional experiments}
In this section, we are showing some supplementary results. First, we show results for the TaylorPDENet on downsampled data with the size $32^2$. Second, we will conduct an experiment investigating the relationship between the number of prediction steps and reconstruction and forecasting MSE.

\begin{figure}[!h]
    \centering
    \begin{subfigure}{0.23\textwidth}
        \centering
        \includegraphics[width=\textwidth]{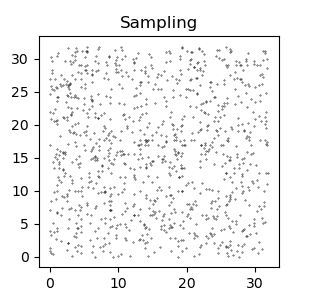}
    \end{subfigure}
    \hfill
    \begin{subfigure}{0.75\textwidth}
        \centering
        \includegraphics[width=\textwidth]{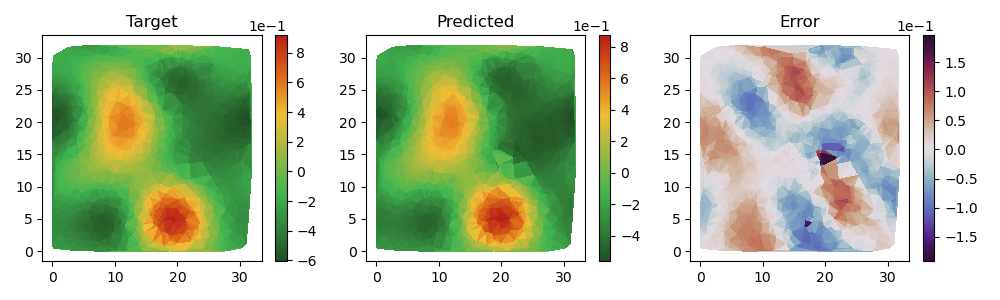}
    \end{subfigure}
\caption{Prediction after $150$ prediction steps for a PDE learned from irregularly spaced data points with only $32^2$ sampled points.}
\label{fig:downsampled_points}
\end{figure}

\subsection{Downsampling}
\begin{figure}[!ht]
    \centering
    \includegraphics[width=\textwidth]{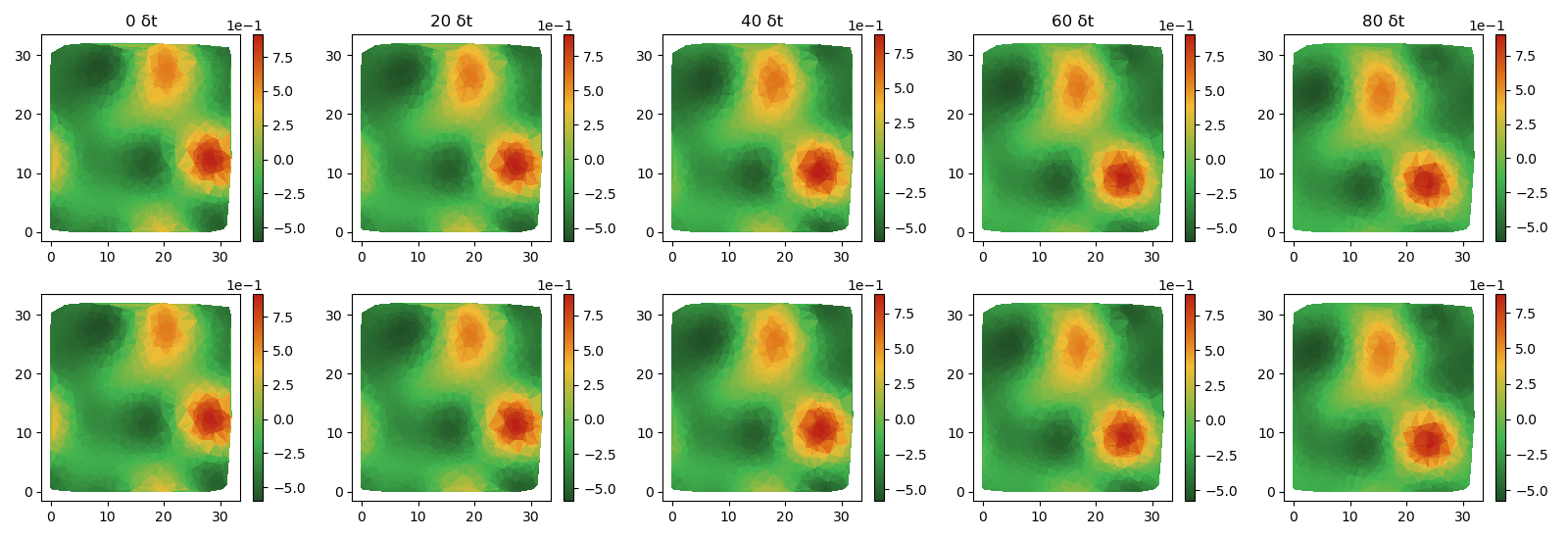}
    \caption{Prediction with TaylorPDENet on $32^2$ downsampled points after an increasing number of prediction steps.}
    \label{fig:32nongrid}
\end{figure}

In \Cref{fig:downsampled_points,fig:32nongrid} results can be seen for the TalorPDENet on $32^2$ points. The effectiveness, even on sparse data, can be seen.

\begin{figure}[!ht]
    \centering
    \includegraphics[width=\textwidth]{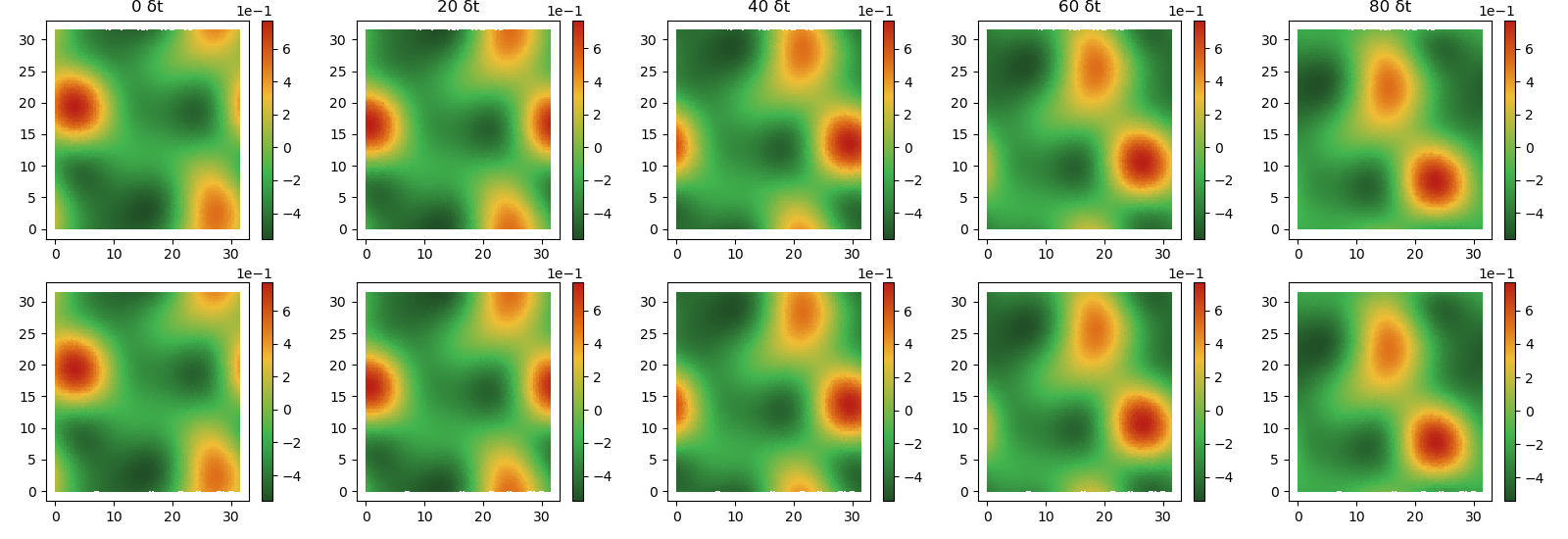}
    \caption{Prediction with TaylorPDENet on $64 \times 64$ points after an increasing number of prediction steps.}
    \label{fig:64grid}
\end{figure}
\subsection{Used prediction steps during training}
\begin{figure}[!ht]
    \centering
    \includegraphics[width=\textwidth]{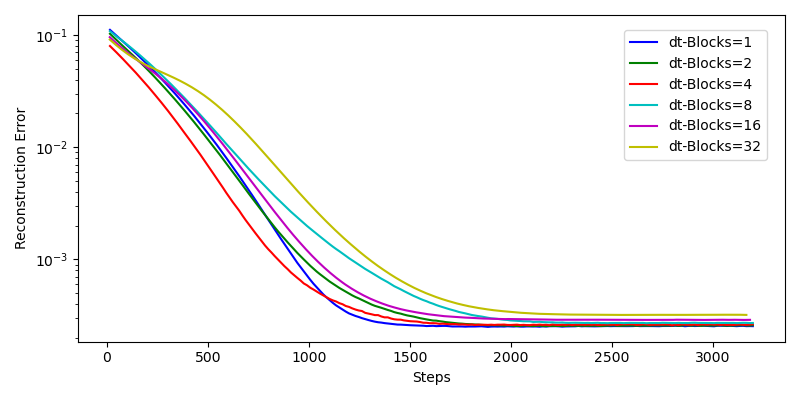}
    \caption{The reconstruction MSE during training. It can be seen that the reconstruction MSE is similar at the end of the training.}
    \label{fig:dt_reconstruction}
\end{figure}

With this experiment, we show the influence of different numbers of prediction steps. In \Cref{fig:dt_reconstruction} the influence of the number of prediction steps during training for the reconstruction can be seen. The reconstruction MSE shows that the final result is in the same order of magnitude. Furthermore, the model is tested with $L=150$ prediction steps as well and it can be seen that the result are very similar. Due to the very similar reconstruction MSE, the prediction is very similar as well.

\begin{table}[!h]
\label{tab:dt_prediction}
\centering
\begin{tabular}{l@{\hskip 0.1in}c@{\hskip 0.1in}c@{\hskip 0.1in}c@{\hskip 0.1in}c@{\hskip 0.1in}c@{\hskip 0.1in}c} 
\toprule
Prediction steps & 1 & 2 & 4 & 8 & 16 & 32\\
\midrule
Forecasting MSE ($10^{-4}$) & $1.07$ & $1.07$ & $1.07$ & $1.06$ & $1.04$ & $1.02$ \\
\bottomrule
\end{tabular}
\caption{The forecasting MSE for $L=150$ prediction steps. The model was trained with the indicated number of prediction steps on non-grid data.}
\end{table}

%
%
%
\bibliographystyle{splncs04}
\bibliography{bibliography}